\pdfoutput=1

\documentclass[11pt]{article}

\usepackage[preprint]{acl}

\usepackage{times}
\usepackage{latexsym}

\usepackage[T1]{fontenc}

\usepackage[utf8]{inputenc}

\usepackage{microtype}

\usepackage{inconsolata}

\usepackage{graphicx}

\usepackage{arydshln}
\usepackage{CJK} 
\usepackage{tcolorbox} 
\usepackage{xcolor} 
\usepackage{graphicx}
\usepackage{algorithm}
\usepackage{algorithmic}

\usepackage{amsmath,bm}
\usepackage{enumitem}
\usepackage{booktabs}
\usepackage{amsfonts,amssymb}
\usepackage{multirow}
\usepackage{arydshln}
\usepackage{adjustbox}
\usepackage{color}
\usepackage{colortbl}
\usepackage{pifont}
\usepackage{subcaption}
\usepackage{soul}
\title{Improving Multilingual Social Media Insights: \\Aspect-based Comment Analysis}

\author{Longyin Zhang, Bowei Zou, and Ai Ti Aw \\
  Institute for Infocomm Research, A*STAR, Singapore \\
  \texttt{\{zhang\_longyin,zou\_bowei,aaiti\}@i2r.a-star.edu.sg} \\}

\begin{document}
\maketitle
\begin{abstract}
The inherent nature of social media posts, characterized by the freedom of language use with a disjointed array of diverse opinions and topics, poses significant challenges to downstream NLP tasks such as comment clustering, comment summarization, and social media opinion analysis. To address this, we propose a granular level of identifying and generating aspect terms from individual comments to guide model attention. Specifically, we leverage multilingual large language models with supervised fine-tuning for comment aspect term generation (CAT{\small $\mathcal{G}$}), further aligning the model's predictions with human expectations through DPO. We demonstrate the effectiveness of our method in enhancing the comprehension of social media discourse on two NLP tasks. Moreover, this paper contributes the first multilingual CAT{\small $\mathcal{G}$} test set on English, Chinese, Malay, and Bahasa Indonesian. As LLM capabilities vary among languages, this test set allows for a comparative analysis of performance across languages with varying levels of LLM proficiency.
\end{abstract}

\renewcommand{\thefootnote}{}%
\footnotetext{This paper was sent to ACL ARR 2024 December for peer-reviewing, with Overall Assessments of 3 by three anonymous reviewers and 4 by Meta Review.}
\renewcommand{\thefootnote}{\arabic{footnote}}

\section{Introduction}
The rapid evolution of mobile technology and the pervasive influence of social media have led to an exponential increase in the dissemination of online news. This phenomenon has spurred significant interest within the research community, particularly in the field of social media text analysis.
Previous studies have explored various facts of this domain, including sensitive comment detection~\cite{pavlopoulos2017deep,chowdhury2020multi,moldovan2022users,sousa2022evaluating}, public opinion mining~\cite{pecar2018towards,gao2019abstractive,yang2019making,huang2023abstractive}, and comment cluster summarization~\cite{aker2016automatic,llewellyn2016improving,barker2016s,vzagar2021unsupervised,zhang2024}. 
\begin{figure}[t]
  \vspace*{5pt}
  \begin{small}
  \begin{center}
  \includegraphics[scale=0.28]{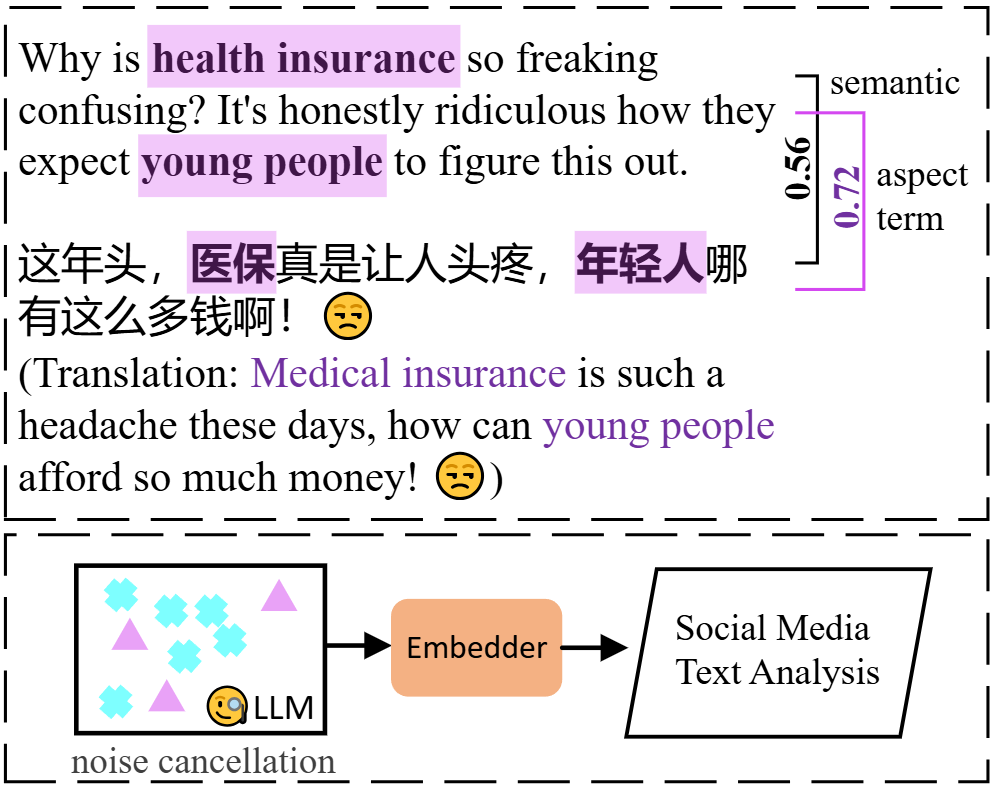}
  \caption{\label{fig:img0} Multilingual social media comments suffer from heavy noise. We harness open-source LLMs to resolve the dilemma by extracting comment aspect terms, effectively acting as a form of noise cancellation, to enhance downstream social media analyses.}
  \end{center}
  \end{small}
\end{figure}

In this paper, we look into the core semantic elements of news comments and focus on Comment Aspect Term Generation (\textbf{CAT{\small $\mathcal{G}$}}) in a multilingual scenario. Social media comments usually suffer from informal expression style with heavy noise and different expression habits from various languages and cultures. As shown in Figure~\ref{fig:img0}, the noise and language mismatch within online comments can result in a relatively low semantic similarity value between comment embeddings~\cite{reimers-gurevych-2020-making} that express similar opinions, which makes downstream tasks like comment clustering and opinion mining very challenging.
In this context, detecting aspect terms from individual comments as references offers a simple yet effective way to mitigate such issues, to capture reliable term-level correlations among comments.
Recognizing the cross-lingual nature of global social media texts and considering the limited research on multilingual aspect term generation, this work lands on CAT{\small $\mathcal{G}$} modeling and corpus construction. 

In summary, our contributions are three-fold:
\begin{itemize}[leftmargin=0.4cm]
\item We introduce the first multilingual CAT{\small $\mathcal{G}$} test data, encompassing language of English (\textbf{EN}), Chinese (\textbf{CN}), Malay (\textbf{MS}), and Bahasa Indonesian (\textbf{ID}). Furthermore, to support advancements in CAT{\small $\mathcal{G}$} and broader social media text analysis, we contribute a manually annotated resource for model fine-tuning to the research community.
\item We continue fine-tune large language models for specialized CAT generation capabilities and further align the distribution of model outputs with that of expert annotations through Direct Preference Optimization (DPO)~\cite{rafailov2024direct}, ensuring user-preferred CAT{\small $\mathcal{G}$}.
\item We integrate the proposed CAT{\small $\mathcal{G}$} system into the downstream task of monolingual and cross-lingual comment clustering (\textbf{ComC}). 
The observed significant performance improvements underscore the value and impact of our work in improving social media text analysis. All codes and data will be publicly available upon email request (CC BY-NC-SA 4.0).
\end{itemize}

\section{Resource \& Approach}
\subsection{Multilingual CAT Annotation}
This research aims to establish a robust framework for identifying and analyzing the central targets of opinions within noisy comment texts.
Building on the work of~\citet{zhang2024}, we define CATs as the primary object(s) that serve as the central focuses of a comment’s opinion, whether expressed explicitly or implicitly.
To ensure consistency in annotation, we strictly adhere to the guidelines outlined by~\cite{zhang2024} with an additional restriction: for lengthy comments that may contain multiple CATs, annotators should prioritize and annotate the five most important aspect terms.

\noindent\textbf{Statistics.} 
Following the annotation guidelines, a professional data annotation team (Asiastar, Singapore) is commissioned to annotate comments in multiple languages at a rate of USD 0.23 per comment for Chinese and Malay, and USD 0.30 for Bahasa Indonesian. Annotations for English comments are conducted in-house by 3 experienced data researchers.
The dataset comprises 5,357 comments from the Reddit forum and the New York Times 2017 dataset\footnote{\url{https://www.kaggle.com/datasets/aashita/nyt-comments}}. For research purposes, we randomly divide the comments into a Fine-tuning and Test dataset, as shown in Table~\ref{table:0}.
\begin{table}[t]
 \small
 \centering
\begin{tabular}{l|ccccc} 
\textbf{Dataset} & \textbf{All} & \textbf{EN} & \textbf{CN} & \textbf{MS} & \textbf{ID} \\
 \hline
Fine-tuning & 2,357 & 809 & 693 & 524 & 331  \\
Test & 3,000 & 1,223 & 814 & 576 & 387  \\
\hline
\end{tabular}
\caption{Manually annotated multilingual CAT corpus.}
\label{table:0}
\end{table}
\begin{table*}[t]
 \small
 \centering
\begin{tabular}{p{2.3cm}|p{4mm} p{4mm} p{5.5mm}|p{4mm} p{4mm} p{5.5mm}|p{4mm} p{4mm} p{5.5mm}|p{4mm} p{4mm} p{5.5mm}|p{4mm} p{4mm} p{5.5mm}}
  \multirow{2}{*}{ \textbf{Method}} & \multicolumn{3}{c|}{\textbf{Overall}} & \multicolumn{3}{c|}{\textbf{EN}}  & \multicolumn{3}{c|}{\textbf{CN}}  & \multicolumn{3}{c|}{\textbf{ID}}  & \multicolumn{3}{c}{\textbf{MS}} \\
  & P & R & F1 & P & R & F1 & P & R & F1 & P & R & F1 & P & R & F1 \\
\hline
SeaLion2 & 19.5 & 39.0 & 26.0 & 25.1 & 40.6 & 31.0 & 18.6 & 29.9 & 22.9 & 13.7 & 44.2 & 21.0 & 18.3 & 42.9 & 25.7 \\
SeaLion2$^\dagger$ & 21.9 & 45.9 & 29.7 & 26.4 & 40.7 & 32.0 & 26.4 & 61.3 & 36.9 & 13.3 & 41.1 & 20.1 & 18.6 & 41.1 & 25.7 \\
SeaLion2DPO$^\dagger$ & 22.4 & 46.6 & 30.3 & 27.0 & 41.6 & 32.7 & 27.4 & 62.3 & 38.0 & 13.5 & 41.1 & 20.3 & 18.8 & 41.6 & 25.9 \\
SeaLion2DPO$^\dagger\spadesuit$ & 22.8 & 46.7 & 30.6 & 27.4 & 41.9 & 33.1 & 27.4 & 62.7 & 38.1 & 13.6 & 40.0 & 20.3 & 19.3 & 41.4 & 26.3 \\
\hline
SeaLLM2 & 7.5 & 14.7 & 9.9 & 7.6 & 13.2 & 9.6 & 8.5 & 17.2 & 11.4 & 5.9 & 14.3 & 8.3 & 7.4 & 14.6 & 9.8 \\
SeaLLM2$^\dagger$ & 23.1 & 47.6 & 31.1 & 24.4 & 38.4 & 29.8 & 29.2 & 70.9 & 41.3 & 14.9 & 43.6 & 22.2 & 20.3 & 40.4 & 27.0 \\
SeaLLM2DPO$^\dagger$ & 26.6 & 47.8 & \underline{34.2} & 33.1 & 45.4 & \underline{38.3} & 31.0 & 63.9 & \underline{41.7} & 17.0 & 43.8 & \underline{24.5} & 21.4 & 38.1 & \underline{27.4} \\
SeaLLM2DPO$^\dagger\spadesuit$ & 27.2 & 45.7 & 34.1 & 34.0 & 43.3 & 38.1 & 32.0 & 61.9 & \textbf{42.2} & 16.3 & 40.3 & 23.2 & 22.1 & 36.5 & \textbf{27.6} \\
\hline
GPT4 & 30.0 & 40.9 & \textbf{34.6} & 36.0 & 41.3 & \textbf{38.5} & 31.9 & 52.2 & 39.6 & 25.4 & 44.2 & \textbf{32.4} & 22.9 & 28.6 & 25.4 \\
\hline
\end{tabular}
\caption{The CAT{\small $\mathcal{G}$} results. ``$\dagger$'' denotes fine-tuning LLMs using the data from GPT4. ``$\spadesuit$'' denotes using a prompt engineering strategy to limit the number of CATs generated. Best results are bolded; second-best are underlined.}
\label{table:2}
\end{table*}

\subsection{CAT{\small $\mathcal{G}$}}\label{cc_sub_ssss}
Recent advancements in NLP have been significantly driven by the emergence of LLMs, with numerous open-source LLMs becoming available. Given the focus on Southeast Asian (SEA) languages in this paper, we take SeaLLM-v2~\cite{nguyen2023seallms} and SeaLion-v2~\cite{sea_lion_2024} as our base models, which are continue-pretrained from Mistral-7B~\cite{jiang2023mistral} and Llama3-8B~\cite{dubey2024llama} and tailored for SEA languages. For supervised fine-tuning (SFT), we acquire the tuning data $D$ by asking GPT4 to generate CATs for multilingual comments, obtaining 5,906 instances with 10\% for validation\footnote{The comments are sourced from the Reddit forum and the New York Times 2017 dataset, with \textbf{no overlap} with the corpus in Table~\ref{table:0}. Additionally, the prompts used for GPT4 requests are provided in Appendix~\ref{sec:appendix3}.}. On this basis, we fine-tune the two small language models with carefully designed instructions (Appendix~\ref{sec:appendix2}) to find an optimal set of parameters $\tau$ as Eq.~(\ref{eq1}), obtaining $\pi_{\tau}$ with foundational CAT{\small $\mathcal{G}$} capabilities. 
\begin{equation}\label{eq1}
\tau = \underset{\tau}{\text{argmin}} \left( \mathbb{E}_{(x,y) \sim D} \left[ L(\pi(x; \tau), y) \right] \right)
\end{equation}

Recognizing that our fine-tuned model is susceptible to GPT-knowledge bias which may be inconsistent with the predefined guidelines and human understanding, we propose incorporating human feedback to generate more user-preferred CATs. Drawing inspiration from~\citet{rafailov2024direct}, who demonstrate the efficacy of DPO in RLHF through a simplified classification loss, we further fine-tune the $\pi_{\tau}$ model on our manual fine-tuning set as Eq.~(\ref{eq2}) to get a parameterized policy $\pi_\theta$.
\begin{align}\label{eq2}
    L(\pi_{\theta}; \pi_{\tau})\!&=\!-\mathbb{E}_{(c, z, z')\sim D'} \bigg[\!\log \sigma \bigg( \beta \log \frac{\pi_{\theta}(z | c)}{\pi_{\tau}(z | c)}  \nonumber \\
    &\quad - \beta \log \frac{\pi_{\theta}(z' | c)}{\pi_{\tau}(z' | c)} \bigg) \bigg]
\end{align}
where $z$ and $z'$ denote the preferred (human annotation) and rejected (GPT4) CATs respectively and $c$ denotes the prompt and comment context. In this way, the resulting model directly optimizes the language model to adhere to user preferences without complicated reward modeling.

\subsection{CAT{\small $\mathcal{G}$}-augmented ComC}\label{cc_sub_ss}
Our research contributes to social media text analysis by providing downstream tasks with a more accurate and refined representation of user comments. To showcase the practical implications of our work, we integrate our CAT{\small $\mathcal{G}$} capabilities into the downstream mono- and cross-lingual ComC tasks~\cite{ma2012topic,llewellyn2016improving,weisser2020clustering}.
While recent work~\cite{zhang2024} proposes a dynamic comment clustering algorithm to mitigate noise through the meticulous design of inner-cluster sentence selection and global cluster ranking, their system relies solely on semantic representations without considering the more specific CAT features. This omission can limit the granularity and precision of comment cluster formation. 

We break through the above limitation by introducing two improvements. On the one hand, we augment the original semantic comment representation by incorporating CAT features through straightforward vector concatenation. This integration allows for a more nuanced representation that captures both the overall semantic and the specific targets of opinions within each comment.
On the other hand, based on our CAT{\small $\mathcal{G}$} results, we classify the comments that do not contain CATs as the \texttt{Trivial} category to eliminate their negative impact on the clustering algorithm.

\section{Experiments}
We calculate the distance between model-generated CATs and the ground truth as the CAT{\small $\mathcal{G}$} performance. Considering the significant variation in language, we set a cosine similarity threshold of 0.7 to define a successful match between two CATs. We report the Precision (\textbf{P}), Recall (\textbf{R}), and \textbf{F1} scores of successful matches as performance. 
For ComC, we follow~\cite{zhang2024} to report the Normalized Mutual Information (\textbf{NMI}) between generated cluster labels and the ground truth. Please refer to Appendix~\ref{sec:appendix1} for more details of system settings.
\begin{figure}[t]
  \begin{small}
  \begin{center}
  \includegraphics[scale=0.63]{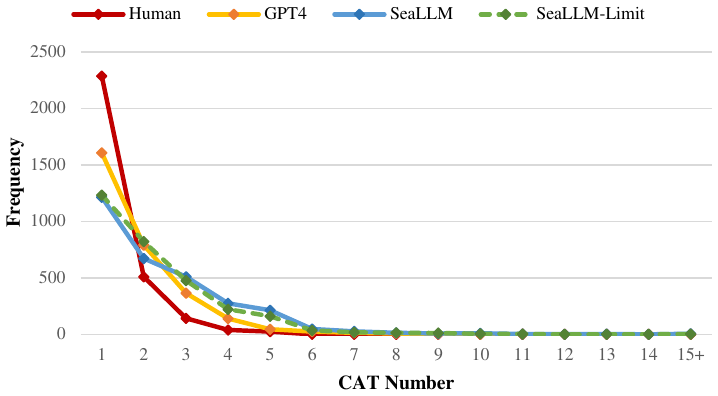}
  \caption{\label{fig:img1} Distribution of Human\&LLM labeled CATs.}
  \end{center}
  \end{small}
\end{figure}

\subsection{Results and Discussion}\label{CATGR}
\paragraph{CAT{\small $\mathcal{G}$} results.}
Table~\ref{table:2} reports the CAT{\small $\mathcal{G}$} performance of GPT4, as well as our fine-tuned SeaLLM-7B~\cite{nguyen2023seallms} and SeaLion-8B~\cite{sea_lion_2024} models.
Firstly, comparing the models before and after fine-tuning (\textbf{$\dagger$}) we find that using GPT4 data can improve open-source small LMs on CAT{\small $\mathcal{G}$}, yielding certain performance improvements. 
Secondly, applying the DPO strategy can further improve the overall performance of the two LLMs.
Besides, the impact of DPO varies across different models and languages, with notable improvements observed for SeaLLM2$^\dagger$ on EN and ID, and for SeaLion2$^\dagger$ on CN.
Thirdly, the relatively high recall scores of our fine-tuned LLMs suggest that the models are prone to over-generating positive predictions. In this case, we further experiment with prompt engineering ($\spadesuit$) to instruct the LLMs to generate fewer CATs. The results show this action works for SeaLLM, resulting in more reliable results with higher precision.
In particular, compared to GPT-4, the final SeaLLM model demonstrates significantly better performance in Chinese and Malay cases, while achieving comparable results in English cases.
Nevertheless, both GPT4 and our fine-tuned models exhibit relatively low performance, indicating the challenges of this task, which will be further discussed in the following.

\paragraph{Distribution of CAT{\small $\mathcal{G}$}.}\label{analysis1}
As mentioned above, the existing results reveal a significant disparity between the current CAT generation and the ground truth, suggesting that while carefully designed prompts can guide LLM behavior, they cannot accurately replicate the nuanced distribution of human-annotated CATs. In this status quo, we conduct a comparative analysis of CAT number distributions to investigate this discrepancy further, as shown in Figure~\ref{fig:img1}. The results highlight a clear pattern: while human annotators typically assign fewer than three CATs per comment, both GPT-4 and our fine-tuned models frequently exceed this threshold. This mismatch reveals the limitations of the prompt-guided generative models in achieving distributional alignment between LLMs and human annotators.
Moreover, the dotted line in Figure~\ref{fig:img1} shows that although our experiment on prompt engineering can slightly influence the distribution of CAT{\small $\mathcal{G}$} (e.g., the prompt-limited system identifies 133 more comments with $\leq 3$ CATs than the baseline), the results ($\spadesuit$) in Table~\ref{table:2} show they fail to resolve the underlying discrepancy fully.
\begin{table}[t]
 \small
 \centering
\begin{tabular}{|p{7.2cm}|} 
 \hline
 \textbf{Comment 1 (Chinese)} \\
\begin{CJK}{UTF8}{gbsn}早点研究出新冠特效药，恢复正常吧！上次回国还是2018年，四年了！\end{CJK}[\textit{I hope we can find a special medicine for COVID-19 soon and get back to normal! The last time I returned home was in 2018, four years ago!}] \\
\textbf{SeaLLM2$^\dagger$}: \begin{CJK}{UTF8}{gbsn}疫情控制, 新冠疫情, 旅行限制, 疫苗\end{CJK} [\textit{epidemic control, COVID-19, travel restrictions, vaccines}]\\
\textbf{SeaLLM2DPO$^\dagger$}: \begin{CJK}{UTF8}{gbsn}回国时间, 新冠特效药\end{CJK} [\textit{return time, special medicine for COVID-19}] \\
\textbf{Human}: \begin{CJK}{UTF8}{gbsn}新冠特效药\end{CJK} [\textit{special medicine for COVID-19}] \\
\hline
\textbf{Comment 2 (Indonesian)} \\
 Laris manis nih jualan senjatanya [\textit{Weapons are selling well}] \\
 \textbf{SeaLLM2$^\dagger$}: Laris manis, jualan senjatanya [\textit{Selling like hot cakes, the selling of weapons}] \\
 \textbf{SeaLLM2DPO$^\dagger$}: NA \\
 \textbf{Human}: NA \\
\hline
\end{tabular}
\caption{Example CAT{\small $\mathcal{G}$} results in Indonesian and Chinese, with corresponding English translations provided.}
\label{table:ana}
\end{table}

\paragraph{Case study.}\label{case_study}
Table~\ref{table:ana} presents two examples with CATs annotated by our system and human annotators. 
In the first example, although the fine-tuned model successfully generated the related CAT of ``\begin{CJK}{UTF8}{gbsn}疫苗\end{CJK} [vaccines]'', it deduces more redundant information like ``\begin{CJK}{UTF8}{gbsn}旅行限制\end{CJK} [travel restrictions]'', which poses a greater risk to the reliability of model outputs. In contrast, the model with DPO applied shows better consistency.
In the second example, the sentence states the facts but lacks the expression of opinions, leading to an ``NA'' tag by annotators. The system results show that the model without DPO training struggles to identify user preferences, a challenge effectively addressed with human feedback applied through DPO. 

\paragraph{CAT{\small $\mathcal{G}$} + ComC.}
Previous research~\cite{zhang2024} proposes the dynamic clustering algorithm (DyClu) tailored for noisy online comments and publishes an English ComC test corpus. In this work, we explore applying CAT{\small $\mathcal{G}$} to ComC to estimate the value of CATs for social media text analysis. 
Building on existing research on comment analysis, we contribute a cross-lingual ComC test set. The dataset is constructed in two stages: first, comments listed in Table \ref{table:0} are grouped according to manually assigned article clusters. Second, comments within each article cluster are further grouped based on human-annotated CATs.
To adapt the DyClu method to the cross-lingual scenario, we employ the multilingual sentence-BERT~\cite{reimers-gurevych-2020-making} for comment representation.
The results in Table~\ref{table:3} show that incorporating our fine-tuned CAT{\small $\mathcal{G}$} model into ComC can significantly improve the overall performance (+2.54 NMI), indicating the significance of CAT{\small $\mathcal{G}$} in social media text analysis. Appendix~\ref{sec:appendix_comc} details the annotation process and case study.
\begin{table}[t]
 \small
 \centering
\begin{tabular}{rl|ccc} 
 ~& \textbf{Method} & \textbf{NMI} \\ 
 \hline
\multirow{2}{*}{M{\scriptsize ONO}L}
&DyClu~\shortcite{zhang2024} & 33.87 \\ 
&DyClu$\P$ & \textbf{34.41} \\ 
\hline
\multirow{2}{*}{C{\scriptsize ROSS}L}
&DyClu~\shortcite{zhang2024} & 40.41 \\ 
&DyClu$\P$ & \textbf{42.95} \\ 
\hline
\end{tabular}
\caption{Mono- and cross-lingual comment clustering performance. ``$\P$'' enhanced with CAT{\small $\mathcal{G}$}.}
\label{table:3}
\end{table}

\section{Conclusion}
This paper investigated the task of CAT{\small $\mathcal{G}$} and its potential to enhance downstream tasks like ComC. 
First, we introduced the first multilingual CAT generation dataset, addressing a critical gap in existing resources.
Second, we established a benchmark for LLM-based CAT generation and further enhanced it with expert knowledge through DPO. 
In addition, we integrated our CAT{\small $\mathcal{G}$} model into downstream mono- and cross-lingual ComC, demonstrating its significance in social media text analysis.
Codes and data will be made available upon email request.

\section{Limitations}
Our work presents three primary limitations. (1) While the DPO method markedly enhances our model's ability to approximate human annotation patterns, a notable performance gap persists. This underscores the pressing need for more advanced methods capable of effectively capturing the subtle nuances and complexities inherent in human annotation distributions.
(2) Unlike ABSA which defines an aspect as a particular attribute, feature, or component of an entity discussed in a text, typically linked to a specific sentiment, our work identifies essential aspect terms to emphasize key attributes or features in public opinion, independent of sentiment. Thus, our research addresses a broader context, concentrating on topic-level or global analysis of comments. Due to space constraints, we could not expatiate this difference in our submission, but we will include it in the final version. (3) Our contributed dataset only considers four languages currently, limiting the research about other SEA languages.

\section{Potential Risks}
This paper presents CAT{\small $\mathcal{G}$}, a model fine-tuned on a limited dataset of noisy online comments and subsequently refined with human feedback via DPO. Due to the small scale of the fine-tuning data and potential biases in the pre-training corpus, the model may exhibit biases. Mitigating these biases requires further investigation into bias detection and removal techniques, developing methods to prevent malicious use of aspect term generation, and exploring privacy-preserving methods for comment analysis.

\section{Acknowledgments}
We thank Nattadaporn Lertcheva, Nabilah Binte Md Johan, Wiwik Karlina, and others for their insightful discussions and contributions to the dataset. 
We thank the reviewers of ACL ARR 2024 December for assigning a high Overall Assessment of 4 (Meta Review) for our submission.

\bibliography{custom}

\appendix
\begin{figure}[t]
    \centering
    \begin{subfigure}{0.45\textwidth}
        \centering
        \includegraphics[width=7cm]{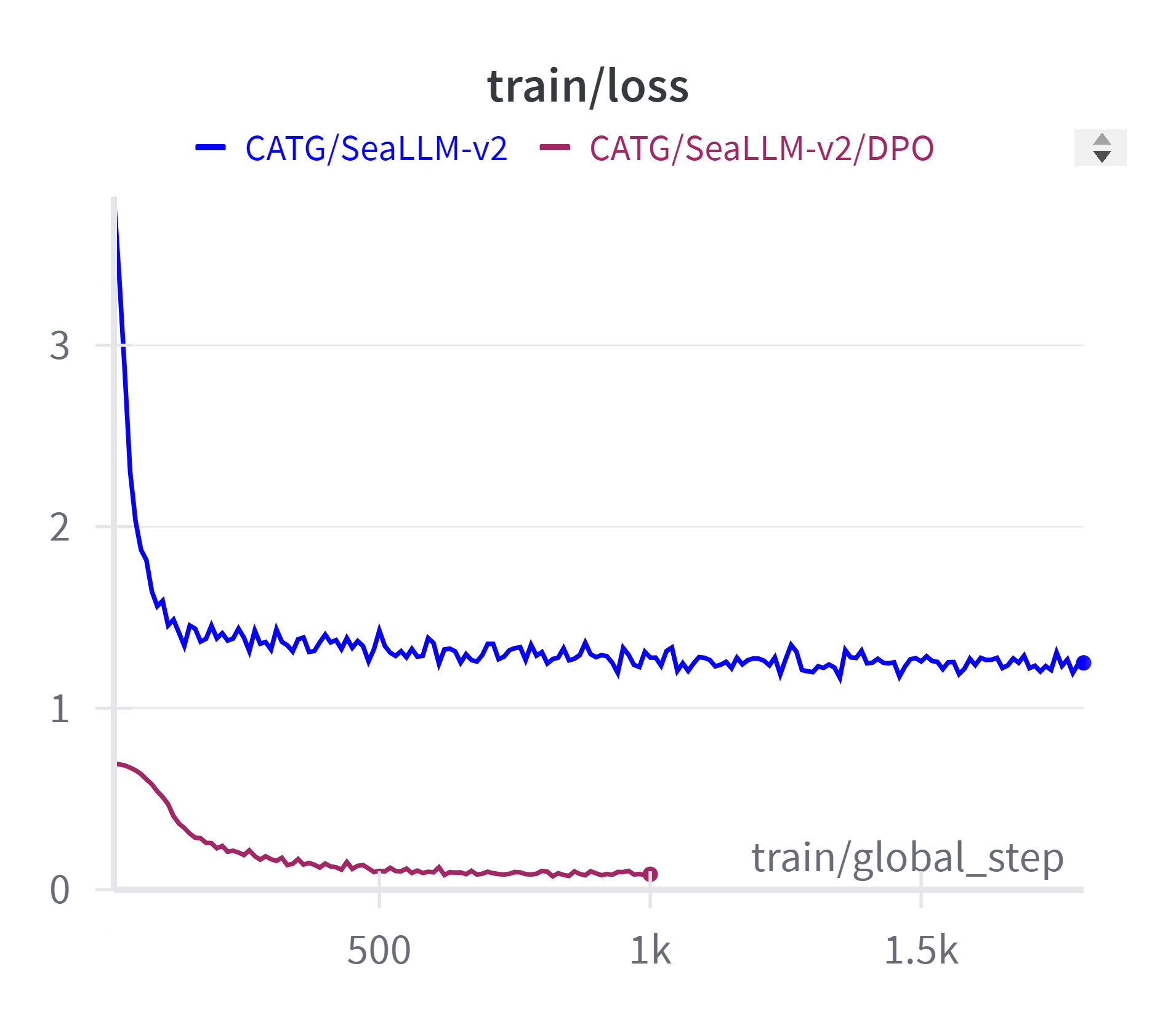}
        \caption{CAT{\small $\mathcal{G}$} Model Training}
        \label{fig:subfig2}
    \end{subfigure}
    \begin{subfigure}{0.45\textwidth}
        \centering
        \includegraphics[width=7cm]{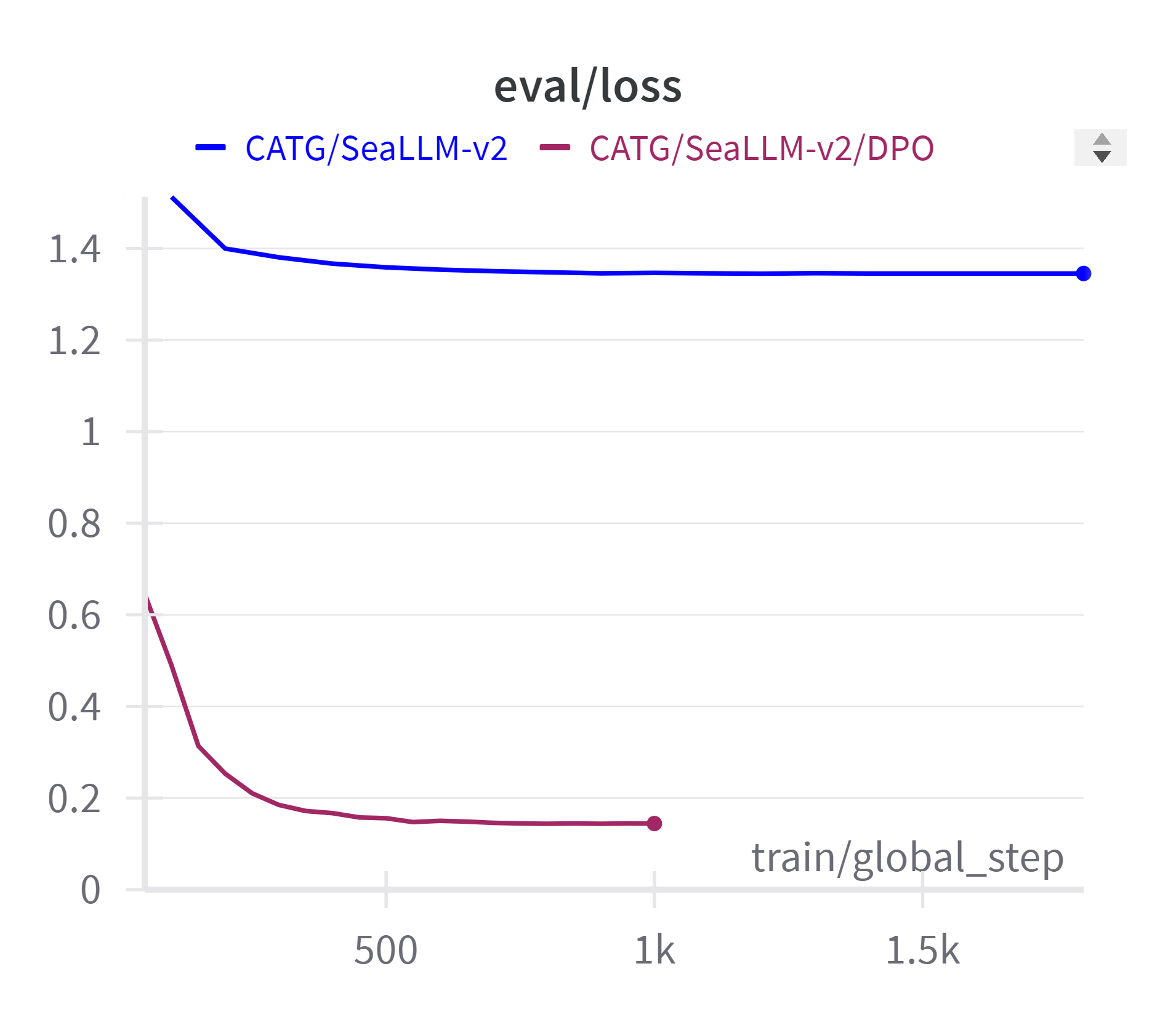} 
        \caption{CAT{\small $\mathcal{G}$} Model Selection}
        \label{fig:subfig1}
    \end{subfigure}
\caption{Model training and validation.}
\label{fig:all}
\end{figure}

\section{System Settings}
\label{sec:appendix1}
\underline{For CAT{\small $\mathcal{G}$}}, we first fine-tuned the LLMs with the 5,906 instances by GPT4 (10\% for validation) over two epochs. Subsequently, we applied DPO for further model fine-tuning, leveraging our annotated CATs in the manual fine-tuning set as accepted answers and GPT4-generated CATs as rejected answers. 
It should be noted that the datasets for SFT and DPO have \textbf{no overlap}.
We report the P, R, and F1 scores of successful matches between generated CATs and ground truth as CAT{\small $\mathcal{G}$} performance. To achieve this, we employed multilingual sentence-BERT~\cite{reimers-gurevych-2020-making} to represent multilingual CATs, and when the cosine similarity between two aspect term embeddings exceeds 0.7 we take it as a successful match. It should be noted that the higher the threshold value, the stricter the evaluation metric and we recommend the following researchers follow this threshold for consistent performance comparison. All systems were implemented using the PyTorch framework and trained on two A40 GPU cards. The model selection process is based on the loss values on the validation set, as shown in Figure~\ref{fig:all}. All results were obtained through multiple experiments, with small fluctuations, which does not change the experimental conclusions. 
\underline{For monolingual ComC}, we fully adopted the system settings outlined in~\cite{zhang2024}. \underline{For cross-lingual ComC}, we updated the monolingual comment representation model to the multilingual model in~\cite{reimers-gurevych-2020-making}. Notably, all the pre-trained models we employ in this work are usable for research purposes.

\section{Test Scale Estimation}\label{analysis2}
\begin{figure}[t]
  \begin{small}
  \begin{center}
  \includegraphics[scale=0.58]{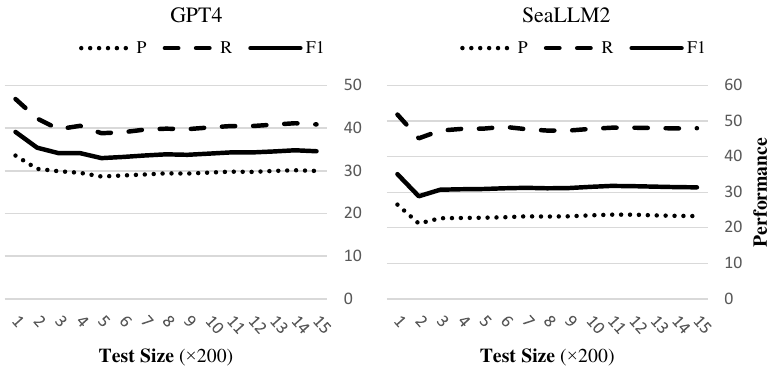}
  \caption{\label{fig:img2} Test data scale analysis.}
  \end{center}
  \end{small}
\end{figure}

Concerns on data contamination in LLM evaluation have been widely acknowledged~\cite{lewkowycz2022beyond}. This concern stems from the possibility of LLMs, particularly those with billions of parameters, memorizing portions of training data, potentially leading to inflated performance metrics on commonly used benchmarks. Therefore, determining an appropriate test size is crucial for reliable model evaluation. 
To investigate the impact of test size on performance stability, we conduct an analysis using varying test sizes. The results in Figure~\ref{fig:img2} show that while performance fluctuates with smaller test sizes, it stabilizes as the size increases, and the final test data is sufficiently large to mitigate the risk of data contamination.

\section{GPT4-based SFT Data Acquisition}
\label{sec:appendix3}
As stated before, we employed 5,906 instances with GPT-annotated CATs to enhance the small LMs with basic CAT{\small $\mathcal{G}$} capabilities through SFT.
Below are the prompts we used to request CATs on EN, CN, MS, and ID from GPT4. It is worth mentioning that since the annotation highlights that each CAT must have corresponding opinion expressions in the comment, we asked GPT to also generate a sentiment label for each comment to better capture the relations between opinions and CATs for more accurate CAT generation.
\begin{tcolorbox}[colback=lightgray!20, colframe=black!75, title=MS]
{\small
Comment Aspect Terms (ATs) mean the main aspects that the comment expresses opinions on, and Emotional Polarity (EP) means the main emotion of the comment. I need you to help me annotate main ATs for each Malay comment (no more than 5 ATs), when no ATs are detected, label ``NA''. For each comment, annotate EP with Negative (N), Positive (P), and Neutral (C). Here shows four annotation examples: Example 1. Bn nak sgt slangor tu bukanya apa..terliur tgk s'gor negeri plg maju hasil negeri billion2 tapi um ... hahaha...ni meols setuju..yelah..selangor kan paling kaya..rizab berbilion billion.. sebab tak dpt nak sakau dr selangor tu yang fed gomen sakit...zaman dedulu masa bn pegang bolehlah sakau sikit [ATs: BN, hasil negara, rizab Selangor, fed gomen | EP: N] Example 2. hahaha...ni meols setuju..yelah..selangor kan paling kaya..rizab berbilion billion.. sebab tak dpt ... kansss..meleleh air liur bn slangor nak merompak duit hasil negeri tapi apakan daya tak dapat.. tapi ada macai desperete dok kait dgn terowong ajaib bagai yg lgsg takde kaitan dgn slangor.. [ATs: rizab selangor, merompak wang | EP: N] Example 3. The best actor goes to... Kesian owner moto. JPJ Dah nampak. takpe kasi chan lepas tu lepas GE claim balik. [ATs: motorcycle owner, JPJ | EP: C] Example 4. Kimarkkkk ko ler jamal tongkol  [ATs: Jamal | EP: N]\textbackslash nAnnotate the following comment ``...'' and return the result as the format of the examples.}
\end{tcolorbox}
\begin{tcolorbox}[colback=lightgray!20, colframe=black!75, title=CN]
{\small
\begin{CJK}{UTF8}{gbsn} 
Comment Aspect Terms (ATs) mean the main aspects that the comment expresses opinions on, and Emotional Polarity (EP) means the main emotion of the comment. I need you to help me annotate main ATs for each Chinese comment (no more than 5 ATs), when no ATs are detected, label ``NA''. For each comment, annotate EP with Negative (N), Positive (P), and Neutral (C). Here are three annotation examples:\textbackslash nAnnotate the following comment ``...'' and return the result as the format of the examples. Example 1. 太假了……我家那里几乎每人都中了只是没人统计而已 [ATs: 新冠统计 | EP: N] 2. 刚才浙江日增100万转到这条新闻成2983起，真的是太不要脸了，还零死亡，现在就我们那里殡仪馆死人都全部放在地上，殡仪馆24小时工作。 [ATs: 网络新闻, 浙江新增病例, 死亡率 | EP: N] Example 3. 呵呵。。。。两声应该明白啥意思 [ATs: NA | EP: C]
\end{CJK}
\textbackslash nAnnotate the following comment ``...'' and return the result as the format of the examples.}
\end{tcolorbox}
\begin{tcolorbox}[colback=lightgray!20, colframe=black!75, title=ID]
{\small
Comment Aspect Terms (ATs) mean the main aspects that the comment expresses opinions on, and Emotional Polarity (EP) means the main emotion of the comment. I need you to help me annotate main ATs for each Indonesian comment (no more than 5 ATs), when no ATs are detected, label ``NA''. For each comment, annotate EP with Negative (N), Positive (P), and Neutral (C). Here shows three annotation examples: Example 1. jumlah nuklir yang dimilik sekutu NATO, China dan Rusia lebih dari cukup untuk bikin bumi kiamat [ATs: NATO, Tiongkok, Rusia, senjata nuklir | EP: N] Example 2. @kampret.strez booster gak ngaruh utk org yg udah kena + di vaksin. itu dari riset empiris dari israel bbrp bulan lalu. gw sih rada skeptis utk ambil booster toh mulai bulan ke 3 antibody udh mulai nurun dan perlu booster lagi dlm 6 bulan.  [ATs: Efek booster, penelitian di Israel | EP: C] Example 3. kalau di Indonesia kebalik yah di beberapa daerah ada yg maksa kapir pake jilbab dgn alasan t0l0l pula macam biar gak digigit nyamuk  [AT: Indonesia, hijab, nyamuk | EP: C]\textbackslash nAnnotate the following comment ``...'' and return the result as the format of the examples.}
\end{tcolorbox}
\begin{tcolorbox}[colback=lightgray!20, colframe=black!75, title=EN]
{\small
Comment Aspect Terms (ATs) mean the main aspects that the comment expresses opinions on, and Emotional Polarity (EP) means the main emotion of the comment. I need you to help me annotate main ATs for each Malay comment (no more than 5 ATs), when no ATs are detected, label ``NA''. For each comment, annotate EP with Negative (N), Positive (P), and Neutral (C). Here shows four annotation examples: Example 1. What it says is that food banks are used to providing support for those in the poorest 10\% of the population but now that segment is creeping up so that more people are needing help. [ATs: food bank, poor singaporeans | EP: N] Example 2. Actually, most [people have savings - in the form of CPF]. [ATs: CPF savings | EP: P] Example 3. And yet every 2 or 3 cars on the road is either bmw or merc [ATs: NA | EP: C]\textbackslash nAnnotate the following comment ``...'' and return the result as the format of the examples.}
\end{tcolorbox}

\section{Instruction Tuning Design}
\label{sec:appendix2}
With the GPT- and human-annotated CATs in hand, we further design the instructions to form the final fine-tuning data.
To encourage the instruction diversity, we prepare a set of 30 brief CAT descriptions (noted as ``CAT-desc'') for random selection to form the instructions:
\begin{tcolorbox}[colback=lightgray!20, colframe=black!75, title=CAT{\small $\mathcal{G}$} Prompt]
{\small
\textbf{\{CAT-desc\}} If no opinion is expressed, the aspect terms should be ``NA''. Annotate the aspect terms of the following comment: ...}
\end{tcolorbox}
\noindent As stated in Subsection~\ref{CATGR}, we perform prompt engineering to limit the CAT generation process, the detailed prompt is as below:
\begin{tcolorbox}[colback=lightgray!20, colframe=black!75, title=CAT{\small $\mathcal{G}$} Limit Prompt]
{\small
\textbf{\{CAT-desc\}} If no opinion is expressed, the aspect terms should be ``NA''. Annotate the following comments with 1 or 2 aspect terms: ...}
\end{tcolorbox}

\section{ComC Case Study}
\label{sec:appendix_comc}
To construct the cross-lingual ComC test set, we followed a multi-step process: (1) group comments (Table \ref{table:0}) by their corresponding article clusters; (2) using the Fast Clustering algorithm, as described in~\cite{zhang2024}, group comments within each article cluster according to manually assigned CAT labels; and (3) manually refine the resulting comment clusters following~\cite{zhang2024}.

To intuitively show the effects of CAT{\small $\mathcal{G}$} on the downstream task of ComC, we illustrate clustering results of DyClu and the CAT{\small $\mathcal{G}$}-enhanced DyClu$\P$ for reference. Considering that a complete article cluster contains a lot of comments, it is difficult to display all comment clusters. Since the DyClu algorithm ensures that the first comment of each cluster represents the cluster centroid, we sampled two comment clusters with the same centroid for analysis, as shown in Table~\ref{table:ana2}. For the results of DyClu, the clustering algorithm only refers to the semantic representation of each comment, making it hard to capture the key points of each comment, so as we can see some unrelated comments are also included in the cluster. On the contrary, the cluster constructed by DyClu$\P$ looks more concise, and all comments are densely focused on ``the current state of Ukraine''.

\begin{table}[h]
 \small
 \centering
\begin{tabular}{|p{7cm}|} 
 \hline
    \textbf{DyClu-source}\\
    {Comment 1}. \begin{CJK}{UTF8}{gbsn}乌呼烂早晚得一命呜呼。 回复\end{CJK} 
    {Comment 2}. dah nk kalah letew..
    {Comment 3}. semoga putin kalah.
    {Comment 4}. bajet sendu..
    {Comment 5}. ermmmm........mcm xde mood nk tgk
    {Comment 6}. nmpk jokowi dimana2..tp ok la atas benda bgus bkn benda tolol
    {Comment 7}. ke rajin claim je ni. tp kalo betol ok la...
    {Comment 8}. lumayannnn pak jokowo
    {Comment 9}. haish lu sapa. stfu je la.
    {Comment 10}. suka2 gw lah... kok lu yg sewot. emang gw mesti satu pendapat dengan lu? kan gak juga gw anti amrik lu fansboy amrik. dari sini aja udah jelaskan? udah gak usah di quote lagi. percuma {Comment 11}. gak nyambung.
    {Comment 12}. aaaahhh...seperti biasa nya, as gertak aja
    {Comment 13}. \begin{CJK}{UTF8}{gbsn}纯属刷存在感。。。 回复\end{CJK} 
    {Comment 14}. \begin{CJK}{UTF8}{gbsn}哈哈哈哈！人才辈出！什么话都敢讲 回复\end{CJK} 
    {Comment 15}. up........
    {Comment 16}. kalah pulak
    {Comment 17}. banyak amoy
    {Comment 18}. ga ngaca dia
    {Comment 19}. mikir
    {Comment 20}. kocak komen nya
    {Comment 21}. rm5.40? hebat-hebat rasa, @uori cukup rasa, eh silap, maggi cukup rasa
    {Comment 22}. adoiii kejam
    {Comment 23}. privet!
    {Comment 24}. this.
    {Comment 25}. wtf.
    {Comment 26}. </blockquote> \\
    \textbf{DyClu-translated}\\
    Comment 1: Sooner or later, Wu Hulan will die.
    Comment 2: It's about to lose, dude.
    Comment 3: Hopefully Putin loses.
    Comment 4: The budget is pathetic.
    Comment 5: Ermmm... I'm not in the mood to watch.
    Comment 6: I see Jokowi everywhere... but okay, it's good stuff, not stupid stuff.
    Comment 7: They just keep claiming things. But if it's true, okay...
    Comment 8: It's alright... Pak Jokowi.
    Comment 9: Who are you? Just shut up.
    Comment 10: I'm free to do whatever I want... Why are you so upset? Do I have to agree with you? I'm not anti-American, you're just an American fanboy. It's clear from this point on. Don't quote me anymore. It's pointless.
    Comment 11: Not relevant.
    Comment 12: Ahh... as usual, just empty threats.
    Comment 13: Purely trying to get attention...
    Comment 14: Hahaha! What talent! They're saying anything they want!
    Comment 15: Up...
    Comment 16: Lost again.
    Comment 17: Lots of girls.
    Comment 18: They don't see themselves.
    Comment 19: Think.
    Comment 20: The comments are hilarious.
    Comment 21: RM5.40? Pretty expensive for what it is, @uori is good enough, wait, I'm wrong, Maggi is good enough.
    Comment 22: Oh dear, that's cruel.
    Comment 23: Privet!
    Comment 24: This.
    Comment 25: What the fuck.
    Comment 26: </blockquote> \\
\hline
    \textbf{DyClu$\P$-source}\\
    {Comment 1}. \begin{CJK}{UTF8}{gbsn}乌呼烂早晚得一命呜呼。 回复\end{CJK} 
    {Comment 2}. dia terkorban demi mempertahankan negara ukraine. pejuang sejati. god bless you.
    {Comment 3}. dia terkorban demi mempertahankan negara ukraine. pejuang sejati. god bless you.
    {Comment 4}. udh ketebak hasilny, nato gk bakal berani nurunin bantuan pasukan paling banter senjata doang. minggu depan ukraina nyerah
    {Comment 5}. dia ada pengalaman kan kat iraq sebab tu nak join bantu ukraine
    help ukraine! \\
    \textbf{DyClu$\P$-translated}\\
    Comment 1: Sooner or later, Wu Hulan will die.
    Comment 2: He sacrificed himself to defend the Ukrainian nation. A true fighter. God bless you.
    Comment 3: He sacrificed himself to defend the Ukrainian nation. A true fighter. God bless you.
    Comment 4: I already know the outcome. NATO won't dare send troops, at best they'll send weapons. Ukraine will surrender next week.
    Comment 5: He has experience in Iraq, that's why he wants to join and help Ukraine. Help Ukraine! \\
\hline
\end{tabular}
\caption{Effects of CAT{\small $\mathcal{G}$} on cross-lingual ComC.}
\label{table:ana2}
\end{table}

\section{Dynamic Clustering}
\label{sec:appendix_alg}
We present the detailed DyClu method in Algorithm~\ref{alg} . This paper employs generated CATs to enhance comment representation for better clustering performance, as highlighted above.

\begin{algorithm}[t]
	\caption{DyClu}
	\label{alg3}
	\begin{algorithmic}
		\STATE \textbf{Input}: $n$ vectorized data points $D$
        \STATE \textbf{Output}: A list of data point clusters $S$
        \STATE \textbf{Initialization}: Initial cluster size: $\gamma$, initial similarity threshold: $\theta$, ceiling threshold: $\theta_{max} = 0.9$
        \STATE \textbf{Begin}
        \FOR{$i$-th data point $d_i$ in $D$}
        \STATE \hl{similarities $\gets$ pairwise\_similarity($d_i$, $D$)}
        \STATE similarities\_top$\gets$ similarities.top-k($\gamma$)
        \STATE $\theta' \gets \theta$
        \WHILE{similarities\_top[-1] > $\theta'$ and $\gamma$ < $n$}
        \STATE $\gamma \gets$ Min($n$, $\gamma + \Delta$)
        \STATE $\theta' \gets$ Min($\sqrt{\text{K}_1 \times (\gamma + \text{K}_2)}$, $\theta_{max}$)
        \STATE similarities\_top$\gets$ similarities.top-k($\gamma$)
        \ENDWHILE
        \STATE $S_i \gets$ []
        \FOR{$s_j$ in similarities\_top}
        \STATE $S_i \gets S_i \cup \{d_j\}$ when $s_j \geq \theta'$
        \ENDFOR
        \STATE $S \gets S \cup \{S_i\}$
        \STATE Calculate the ranking score $R_i$
        \ENDFOR
        \STATE Rank the $n$ clusters in $S$ based on $R_{1\dots n}$
    \STATE \textbf{End}
	\end{algorithmic}
\label{alg}
\end{algorithm}

\section{Supplementary Examples about CAT{\small $\mathcal{G}$} Evaluation}
\label{sec:appendix_catgeve}
Table~\ref{table:ana_add1} illustrates a set of CAT pairs with distances close to the manually defined similarity threshold 0.7 for reference.

\begin{table}[h]
\centering
\small
\renewcommand{\arraystretch}{1.5} 
\begin{tabular}{|p{2.8cm}|p{2.8cm}|c|}
\hline
\textbf{Category A} & \textbf{Category B} & \textbf{Dis.} \\
\hline
cheap labour mindset & cheap labour & 0.77 \\
MC (Medical Certificate) & MC & 0.72 \\
major symptoms & people with major symptoms & 0.78 \\
poverty line & poor line & 0.75 \\
safe distancing & safe distancing on the train & 0.72 \\
welfare improvement & healthcare workers welfare & 0.65 \\
changes to the measures & changes in COVID-19 measures & 0.62 \\
financial aspects & financial support & 0.67 \\
booster shots & booster & 0.68 \\
impact of rising prices on the poor & rising prices & 0.60 \\
\hline
\end{tabular}
\caption{CAT pairs with distances around the artificially set similarity threshold.}
\label{table:ana_add1}
\end{table}

\end{document}